\documentclass[review]{elsarticle}

\usepackage{hyperref}
\usepackage{graphicx}
\usepackage{epstopdf}
\usepackage{CJK}

\usepackage{amsmath}
\usepackage{amssymb}
\usepackage{amsfonts}
\usepackage{mathrsfs}
\usepackage{bm}

\usepackage{booktabs}
\usepackage{multirow}
\usepackage{threeparttable}
\usepackage{tabularx}
\usepackage{makecell}
\usepackage{diagbox}
\usepackage{colortbl}

\usepackage[table,xcdraw,dvipsnames]{xcolor}

\definecolor{myPink}{rgb}{0.9294, 0.0078, 0.5490}
\definecolor{Gray}{gray}{0.92}
\definecolor{my_color}{HTML}{E8F3F1}
\definecolor{my_color1}{HTML}{FFEACE}
\definecolor{my_color2}{HTML}{FBEAFF}
\definecolor{my_color3}{HTML}{FFC1B5}

\usepackage[subfigure]{tocloft}
\usepackage{subfigure}
\usepackage{float}
\usepackage{afterpage}
\usepackage{placeins}

\usepackage[linesnumbered,ruled,vlined]{algorithm2e}

\SetKwComment{Comment}{$\triangleright$\ }{}
\SetKwInput{KwInput}{Input}
\SetKwInput{KwOutput}{Output}
\SetKwInOut{Initialization}{Initialize}

\usepackage{pifont}
\usepackage{microtype}
\usepackage{ragged2e}

\begin{document}
\begin{frontmatter}
	
\title{FreqFLD: Towards All-in-One Facial Landmark Detection via Frequency Modulation}

\author[label1]{Shun Ren}
\ead{renshun@ctgu.edu.cn}

\author[label1]{Kaijie Jin}
\ead{1059657014@qq.com}


\author[label2]{Shengkai Hu\corref{cor1}}
\ead{shengkaihu@stu.zuel.edu.cn}

\author[label3]{Beihang Song \corref{cor1}}
\ead{bhs605906@whu.edu.cn}

\author[label1]{Hang Sun}
\ead{sunhang0418@whu.edu.cn}

\author[label4]{Wenwen Min }
\ead{minwenwen@ynu.edu.cn}

\author[label5]{Youfa Liu}
\ead{liuyfa1991@whu.edu.cn}

\author[label2,label6]{Jun Wan}
\ead{N2409681J@e.ntu.edu.sg}

\cortext[cor1]{Corresponding author: Shengkai Hu and Beihang Song.}

\address[label1]{College of Computer and Information Technology, China Three Gorges University, Yichang 443002, China}
\address[label2]{School of Information Engineering, Zhongnan University of Economics and Law, Wuhan 430073, China}
\address[label3]{National Institute of Natural Hazards, Ministry of Emergency Management of China, Beijing 100085, China}
\address[label4]{School of Information Science and Engineering, Yunnan University, Kunming, Yunnan 650091, China}
\address[label5]{School of Computer Science, Wuhan University, Wuhan 430072, China}
\address[label6]{School of Computer Science and Engineering, Nanyang Technological University, Singapore 639798, Singapore}



\begin{abstract}
Recent progress in deep learning has significantly advanced facial landmark detection.
However, most existing methods process features in a spatial-domain manner under a dataset-specific training paradigm, which overlooks the fact that facial landmark detection is inherently geometry-driven and sensitive to frequency variations, thereby limiting cross-dataset generalization under complex scenarios and hindering the development of a facial landmark detection model. To address this issue, we propose \textbf{FreqFLD}, a \textbf{freq}uency-modulated framework towards All-in-One \textbf{f}acial \textbf{l}andmark \textbf{d}etection. Specifically, FreqFLD introduces a Frequency Modulation Module (FreqMoM) to explicitly induce the frequency prior by decoupling and modulating low- and high-frequency components, which is then injected into subsequent feature modeling to enable balanced modeling of global facial structure and local landmark details. 
Furthermore, FreqFLD employs a Frequency-Modulated Mixture-of-Experts (FreqMoE), with expert selection adaptively conditioned on frequency-modulated priors, enabling flexible modeling of heterogeneous facial landmark patterns under diverse and challenging scenarios. To regularize frequency-consistent modeling under the All-in-One paradigm, we further introduce a Frequency-Consistent Routing (FreqCR) loss, which constrains the routing and assignment of frequency-aware experts to promote balanced expert utilization across diverse facial scenarios, thereby enabling stable expert specialization and achieving robust facial landmark detection. Extensive experiments demonstrate that the proposed FreqFLD achieves comparable performance on popular datasets. 
The code is available at: \href{https://github.com/jkj1059657014/FreqFLD}{https://github.com/jkj1059657014/FreqFLD}.
\end{abstract}

\begin{keyword}
\ Facial Landmark Detection\sep Heatmap regression\sep All-in-One\sep Frequency Learning\sep Multi-Dataset Training
\end{keyword}

\end{frontmatter}


\section{INTRODUCTION}
Facial landmark detection (FLD), also known as face alignment, is a fundamental task in computer vision that aims to localize semantically consistent keypoints (e.g., eyes, nose tip, mouth corners, etc.) for facial geometry and structure. It serves as a critical prerequisite for a wide range of downstream applications, including face recognition~\cite{matsugu2003subject,yoo2015optimized}, expression recognition~\cite{ioannou2005emotion,tang2024facial}, 3D face reconstruction~\cite{feng2018joint,basak20223d} and virtual avatar generation~\cite{xu2023seeavatar,chu2024gpavatar}.

\begin{figure}[t]
	\begin{center}
		\includegraphics[width=0.8\linewidth]{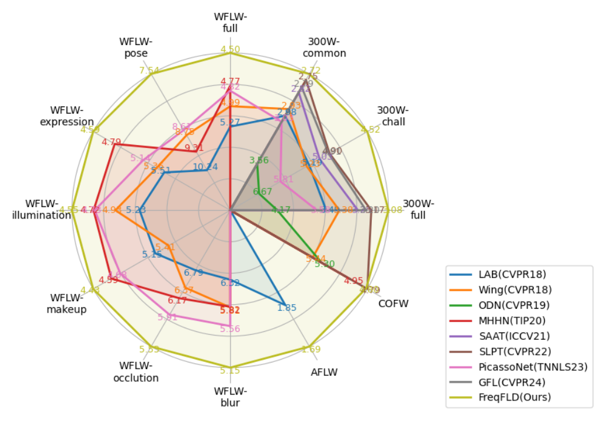}
	\end{center}
    \vspace{-2em}
	\caption{The proposed FreqFLD facilitates unified learning across multiple facial landmark detection datasets with heterogeneous annotation schemes and demonstrates comparable performance.}
	\label{radar}
\end{figure}

With the advancement of deep learning, most existing FLD methods follow a dataset-specific training paradigm~\citep{kowalski2017deep,Wan2019RobustFA,wan2023precise,Wan2025InterpretableFL}, where models are independently optimized for individual benchmarks. 
While such a paradigm has achieved impressive performance on single datasets, it often suffers from limited generalization across datasets and incurs considerable computational and maintenance costs due to repeated training. 
A comparison across various FLD benchmarks indicates that, despite variations in the number of annotated landmarks (68 for 300W, 19 for AFLW, 29 for COFW, and 98 for WFLW), all datasets describe the same fundamental facial geometry. In particular, overlapping semantic landmarks across different datasets span a shared geometric subspace, which is repeatedly learned by dataset-specific models. This structural redundancy suggests that a unified representation can be learned across heterogeneous datasets, thereby motivating a unified architecture for FLD. Moreover, to improve cross-domain generalization, more recently, inspired by advances in low-level image restoration~\citep{ma2025evoir,hu2025clusir,chen2026lovif,hu2026spikerestormer,ClearAIR}, and fine-grained visual understanding~\citep{wan2025fine,Perceive-IR,min2024multimodal}, the All-in-One paradigm has been explored to handle multiple degradations within a single framework. However, extending these paradigms to FLD inevitably exacerbates feature conflicts induced by large variations in facial pose, expression, and occlusion.

More recently, mixture-of-experts (MoE) architectures~\citep{tuzel2016robust,arnaud2019tree,hu2026proto,UniUIR} have been explored within the All-in-One paradigm as an effective mechanism to alleviate feature conflicts by employing multiple specialized experts to model heterogeneous data distributions through adaptive routing.
However, extending existing MoE frameworks to FLD presents non-trivial challenges. On the one hand, traditional individual experts are primarily based on spatial-domain feature modeling, which limits their ability to capture the intrinsic geometric structure of faces.
On the other hand, unconstrained expert routing leads to unstable specialization, resulting in expert imbalance or inconsistent assignment of facial patterns across experts, which further compromises training stability and generalization performance. These limitations are further exacerbated under multi-dataset joint training in the All-in-One paradigm.

Motivated by these observations, we propose FreqFLD, a frequency-modulated All-in-One framework for robust FLD. FreqFLD incorporates a FreqMoM that explicitly decomposes input features into complementary low- and high-frequency components.
By separately modeling low-frequency structural information and high-frequency landmark-sensitive details, FreqMoM constructs frequency-disentangled priors. Furthermore, we introduce a FreqMoE, in which expert routing is adaptively conditioned on frequency-modulated priors.
By guiding expert selection with frequency-aware structural cues, FreqMoE promotes consistent expert specialization across datasets, enabling different experts to capture heterogeneous facial patterns while preserving shared geometric structure, thereby mitigating feature conflicts induced by diverse facial conditions. To further regularize frequency-consistent expert specialization under the All-in-One paradigm, we propose a FreqCR loss.
By constraining expert assignment to remain aligned with frequency-aware structural cues, FreqCR stabilizes expert routing dynamics and prevents expert imbalance during training, leading to more stable optimization and improved cross-dataset generalization. By jointly integrating FreqMoM, FreqMoE, and FreqCR, FreqFLD achieves coherent frequency-aware representation learning and stable expert specialization under heterogeneous facial statistics, leading to comparable FLD performance (Fig.~\ref{radar}). The main contributions of this work are summarized as follows:

(1) We propose \textbf{FreqFLD}, a frequency-modulated All-in-One framework for FLD, which alleviates cross-dataset feature conflicts by integrating frequency-aware representation learning with expert-based modeling. The proposed FreqFLD achieves competitive FLD performance compared with dataset-specific training methods.

(2) We introduce the \textbf{Frequency Modulation Module (FreqMoM)} and the \textbf{Frequency-Modulated Mixture-of-Experts (FreqMoE)}, two complementary components that respectively enable frequency-disentangled feature modeling and frequency-conditioned expert specialization, facilitating robust facial structure modeling under diverse facial conditions.

(3) We propose a \textbf{Frequency-Consistent Routing (FreqCR) loss} to regularize expert routing, promoting stable and frequency-consistent specialization under the All-in-One paradigm.

\section{RELATED WORK} \label{Related Work}
FLD can be traced back to the end of the 19th century. Early approaches relied on handcrafted models, such as Active Shape Models (ASM)~\citep{yan2003face,zheng2008facial}, Constrained Local Models (CLM)~\citep{cristinacce2006feature,cristinacce2008automatic}, and random forest–based methods~\citep{kazemi2014one,feng2014random}, which exhibit limited robustness under unconstrained conditions. With the advent of deep learning, FLD has shifted toward deep learning-based methods~\citep{wan2021robust,wan2024precise,wan2026fgtbt,wan2026universal}, which can be broadly categorized into coordinate regression and heatmap regression paradigms.

\subsection{Coordinate Regression Methods}
Coordinate regression methods formulate facial landmark detection as a direct coordinate prediction problem, where landmark locations are learned by minimizing regression losses between predicted and ground-truth coordinates. This formulation enables seamless integration into end-to-end learning frameworks and avoids the intermediate heatmap representation. Early coordinate regression approaches~\citep{wu2017leveraging} primarily improve robustness through data-driven modeling strategies that leverage dataset-level variations for generalization. In particular, modeling inter- and intra-dataset variations has been shown to facilitate cross-dataset face alignment. To alleviate localization sensitivity, loss re-weighting strategies~\citep{feng2018wing} have been introduced to emphasize small and medium-range errors, thereby enhancing robustness under challenging conditions. Subsequent studies further enhance coordinate regression by explicitly modeling structural dependencies among facial landmarks. Coarse-to-fine frameworks~\citep{gao2020coarse} incorporating landmark-guided self-attention capture global contextual relationships and improve spatial consistency. In addition, graph-based relational modeling~\citep{lin2021structure} has been explored to encode inter-landmark dependencies explicitly, enabling more structured reasoning over facial geometry. More recently, sparse and adaptive interaction mechanisms~\citep{xia2022sparse} have been introduced to efficiently model landmark relationships while reducing computational overhead. Despite these advances, coordinate regression methods remain sensitive to variations in facial geometry and spatial configurations, which limits their robustness in scenarios involving large pose changes, occlusions, or complex expressions.

\subsection{Heatmap Regression Methods}
In contrast to coordinate regression, heatmap regression methods formulate facial landmark detection as a dense spatial prediction problem, where each landmark is represented by a probability heatmap. This formulation provides richer spatial supervision and has therefore become the dominant paradigm, offering improved robustness to pose variations and partial occlusions. Early heatmap-based approaches primarily enhance structural modeling through stacked convolutional architectures. By integrating multi-stage refinement with shape normalization mechanisms, stacked hourglass-style networks~\citep{yang2017stacked} effectively reduce shape variance and improve localization accuracy. Hierarchical ensemble strategies~\citep{zou2019learning} further exploit multi-level representations to implicitly capture inter-landmark relationships and structural consistency. Subsequent studies strengthen robustness by explicitly modeling facial structure and occlusion patterns. Part–whole representations~\citep{zhu2022occlusion} have been introduced to handle severe occlusions by decomposing facial geometry into structured components. In addition, masked representation learning combined with correspondence refinement~\citep{yin2024sce} has been explored to improve landmark localization under missing or corrupted regions. More recently, transformer-based architectures~\citep{dang2025cascaded} have been adopted to capture long-range spatial dependencies and global contextual interactions across landmarks, enabling more holistic modeling of facial geometry. Despite their strong performance, heatmap regression methods largely operate in the spatial domain and rely on dataset-specific supervision, which makes them sensitive to domain shifts and limits their generalization across datasets.


Overall, existing FLD methods mainly focus on dataset-specific optimization and spatial-domain modeling, which limits their generalization across diverse datasets and scenarios.
Motivated by these limitations, we propose a unified framework that incorporates FreqMoM and FreqMoE with adaptive expert routing, further regularized by the FreqCR loss, for robust FLD across diverse datasets under the All-in-One paradigm.

\begin{figure}[t]
\begin{center}
	\includegraphics[width=1\linewidth]{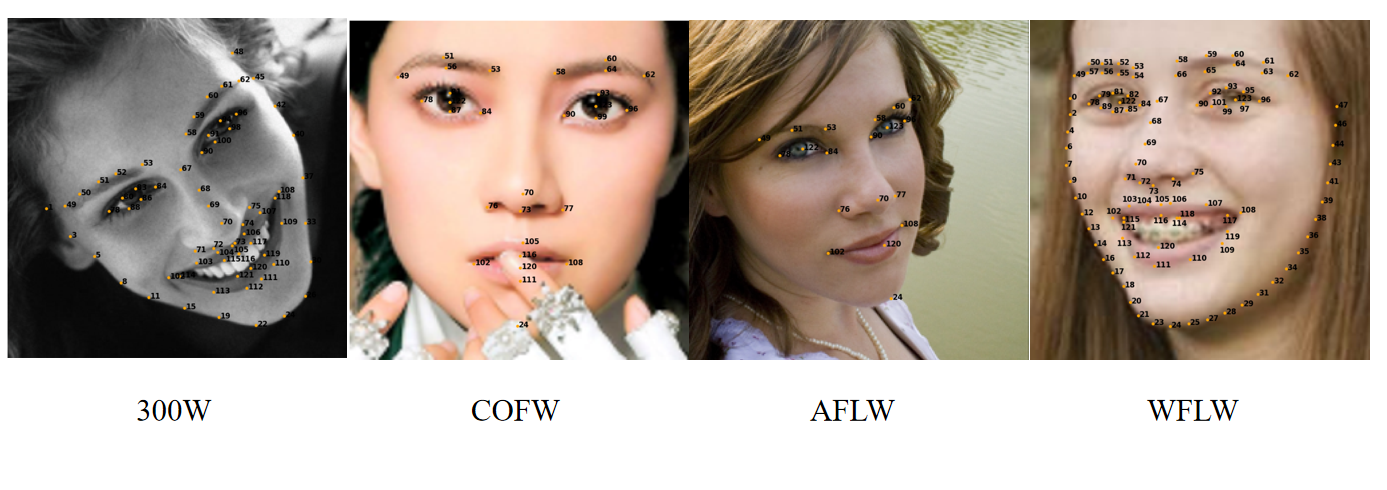}
	\end{center}
    \vspace{-3em}
	\caption{Proposed unified landmark index. A total of 124 unified landmarks are constructed by merging the original landmark annotations from four commonly used facial landmark datasets.}
	\label{UFLD}
\end{figure}

\section{Method}\label{Method}
In this section, we first introduce the unified landmark index in \textbf{Section~\ref{sec:ufld}}, followed by the overall pipeline in \textbf{Section~\ref{sec:Overall}}.
The core components, including the FreqMoM and the FreqMoE, are described in \textbf{Section~\ref{sec:FreqMoM}} and \textbf{Section~\ref{sec:FreqMoE}}, respectively. Finally, the FreqCR loss is introduced in \textbf{Section~\ref{sec:Loss}}.

\subsection{Unified Landmark Index}
\label{sec:ufld}
Different FLD datasets contain different numbers of landmarks, and some landmarks in different datasets correspond to the same semantic information. Hence, to construct a universal landmark detection paradigm, we first define a universal landmark version that aims to contain all landmarks from the popular facial landmark datasets including 300W, WFLW, COFW, and AFLW. Each landmark in the universal landmark version has a specific index for identification. As a result, we establish a set of 124 universal landmarks with well-defined semantics and a unified indexing system for consistent referencing, as shown in Fig.\ref{UFLD}. With the proposed universal landmark definition, FreqFLD can leverage shared landmark annotations across diverse datasets, thereby improving the overall detection accuracy.

\begin{figure*}[t]
	\begin{center}
		\includegraphics[width=1\linewidth]{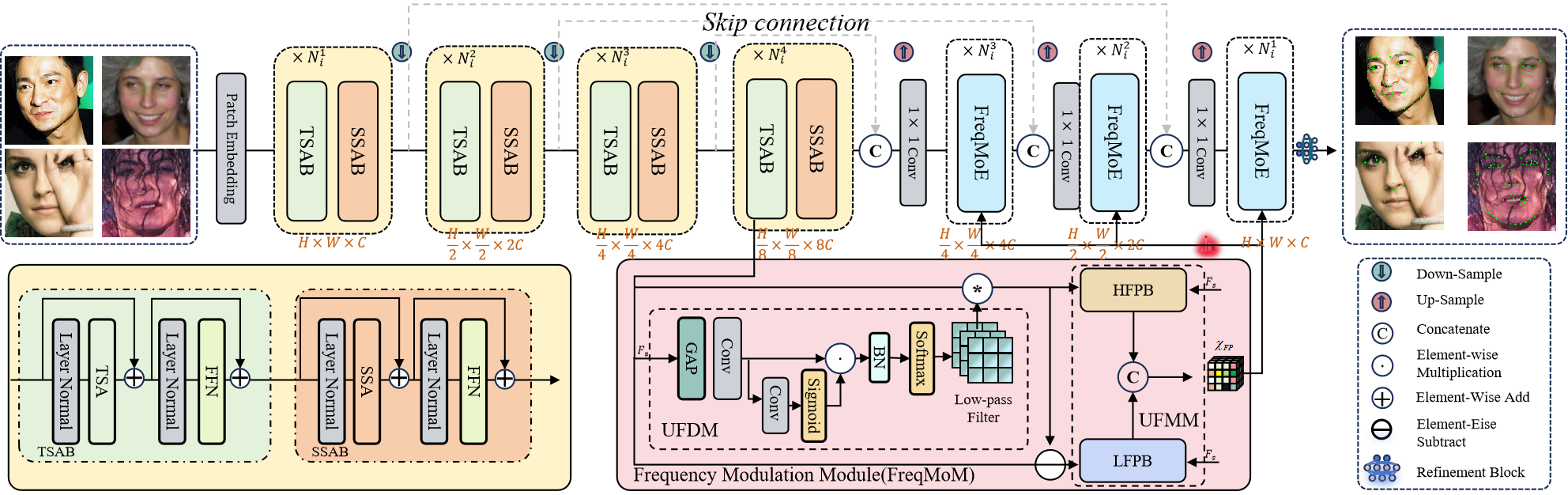}
	\end{center}
    \vspace{-1.5em}
	\caption{The proposed FreqFLD first preprocesses the input image and extracts hierarchical representations through a multi-scale encoder-decoder architecture composed of TSAB and SSAB blocks. At the bottleneck stage, features are decomposed by FreqMoM into high- and low-frequency components, which are further refined via HFPB and LFPB to generate frequency-aware prompts. These prompts guide the decoder through the FreqMoE, enabling dynamic expert activation according to sample complexity. The refined output is subsequently processed by the final prediction head to obtain the unified landmark index.} 
\label{stru}
\end{figure*}

\subsection{Overall Pipeline}
\label{sec:Overall}
We propose FreqFLD (Fig.~\ref{stru}), a frequency-modulated All-in-One framework for robust FLD. Given an input facial image $I \in \mathbb{R}^{H \times W \times 3}$, FreqFLD first applies a $3\times3$ convolutional embedding layer to obtain shallow features $\mathbf{F} \in \mathbb{R}^{B \times C' \times H \times W}$. The embedded features are then processed by a multi-stage encoder--decoder architecture with skip connections for hierarchical feature fusion. To expose geometry-sensitive frequency cues, features $\mathbf{F}_s \in \mathbb{R}^{\frac{H}{8} \times \frac{W}{8} \times 8C}$ at the intermediate stage are processed by the proposed FreqMoM, which decomposes them into complementary low-frequency structural components and high-frequency detail components and produces frequency-modulated priors. These priors are subsequently leveraged in the decoder to guide the FreqMoE, where expert routing is adaptively conditioned on the frequency-aware representations to model heterogeneous facial patterns. After the refinement block aggregation, the features are progressively upsampled to produce dense landmark-aware feature maps 
$\hat{\mathbf{F}} \in \mathbb{R}^{B \times C \times \frac{H}{4} \times \frac{W}{4}}$.
Finally, a lightweight prediction head generates high-resolution landmark heatmaps 
$\hat{\mathbf{Y}} \in \mathbb{R}^{B \times 124 \times \frac{H}{2} \times \frac{W}{2}}$,
where 124 denotes the number of unified facial landmarks.
A FreqCR loss is further introduced to regularize expert assignment during training under the All-in-One  paradigm.

\subsection{Frequency Modulation Module (FreqMoM)}
\label{sec:FreqMoM}
FLD inherently involves heterogeneous frequency characteristics, with low-frequency components modeling global facial geometry and high-frequency components emphasizing fine-grained landmark details. However, existing FLD methods predominantly operate in the spatial domain, without explicitly modeling frequency heterogeneity across regions.
Under the All-in-One setting, this spatial-domain bias amplifies cross-dataset conflicts, weakening fine-grained landmark representations. To address this issue, we propose FreqMoM, which explicitly decomposes intermediate features into complementary low- and high-frequency components and adaptively modulates them to construct frequency-aware features.

\noindent\textbf{Unified Frequency Decoupling Module (UFDM).}
Given an input shallow feature map $\mathbf{F}_s$, FreqMoM first applies global average pooling (GAP) followed by a $1\times1$ convolution to extract a compact global descriptor $\mathbf{F}'_s$. To generate an input-adaptive low-pass filter, $\mathbf{F}'_s$ is further processed by a lightweight gating branch. Specifically, the gating branch predicts a sigmoid-normalized modulation map, which is multiplied with $\mathbf{F}'_s$ to obtain the gated representation $\hat{\mathbf{F}}_s$. The gated representation is then normalized by batch normalization and a softmax operation to produce the adaptive low-pass filter $\mathbf{K}_{\mathrm{l}}$. This process is formulated as:
\begin{equation}
\mathbf{F}'_s
=
\operatorname{Conv}_{1\times1}\!\left(
\mathrm{GAP}(\mathbf{F}_s)
\right),
\end{equation}
\begin{equation}
\hat{\mathbf{F}}_s
= 
\operatorname{Sigmoid}
\big(
\mathcal{G}(\mathbf{F}'_s)
\big) \odot \mathbf{F}'_s,
\end{equation} 
\begin{equation}
\mathbf{K}_{\mathrm{l}}
=
\operatorname{Softmax}
\left(
\mathcal{BN}(\hat{\mathbf{F}}_s)
\right),
\end{equation}
where $\mathcal{G}(\cdot)$ is implemented by a $1\times1$ convolution, $\odot$ denotes element-wise multiplication, $\mathcal{BN}(\cdot)$ denotes batch normalization, and $\mathbf{K}_{\mathrm{l}}$ denotes the learned adaptive low-pass filter, which is then applied to the input feature $\mathbf{F}_s$ through a convolution operation to obtain the low-frequency features. The filtering operation is formulated as:
\begin{equation}
\mathbf{F}_{\mathrm{l}}=
\mathbf{K}_{\mathrm{l}} * \mathbf{F}_{s},
\end{equation}
where $*$ denotes the convolution operation and $\mathbf{F}_{\mathrm{l}}$ denotes the low-frequency features. Furthermore, the high-frequency features are obtained through an element-wise subtraction operation, which can be defined as:
\begin{equation}
\mathbf{F}_{\mathrm{h}} = \mathbf{F}_s - \mathbf{F}_{\mathrm{l}},
\end{equation} where $\mathbf{F}_{\mathrm{h}}$ is complementary to $\mathbf{F}_{\mathrm{l}}$, with low-frequency components capturing global facial geometry and high-frequency components emphasizing fine-grained structural details. Together, they serve as effective priors that guide subsequent feature modeling toward geometrically informative regions within a unified design.

\begin{figure}[t]
\begin{center}
\includegraphics[width=1\linewidth]{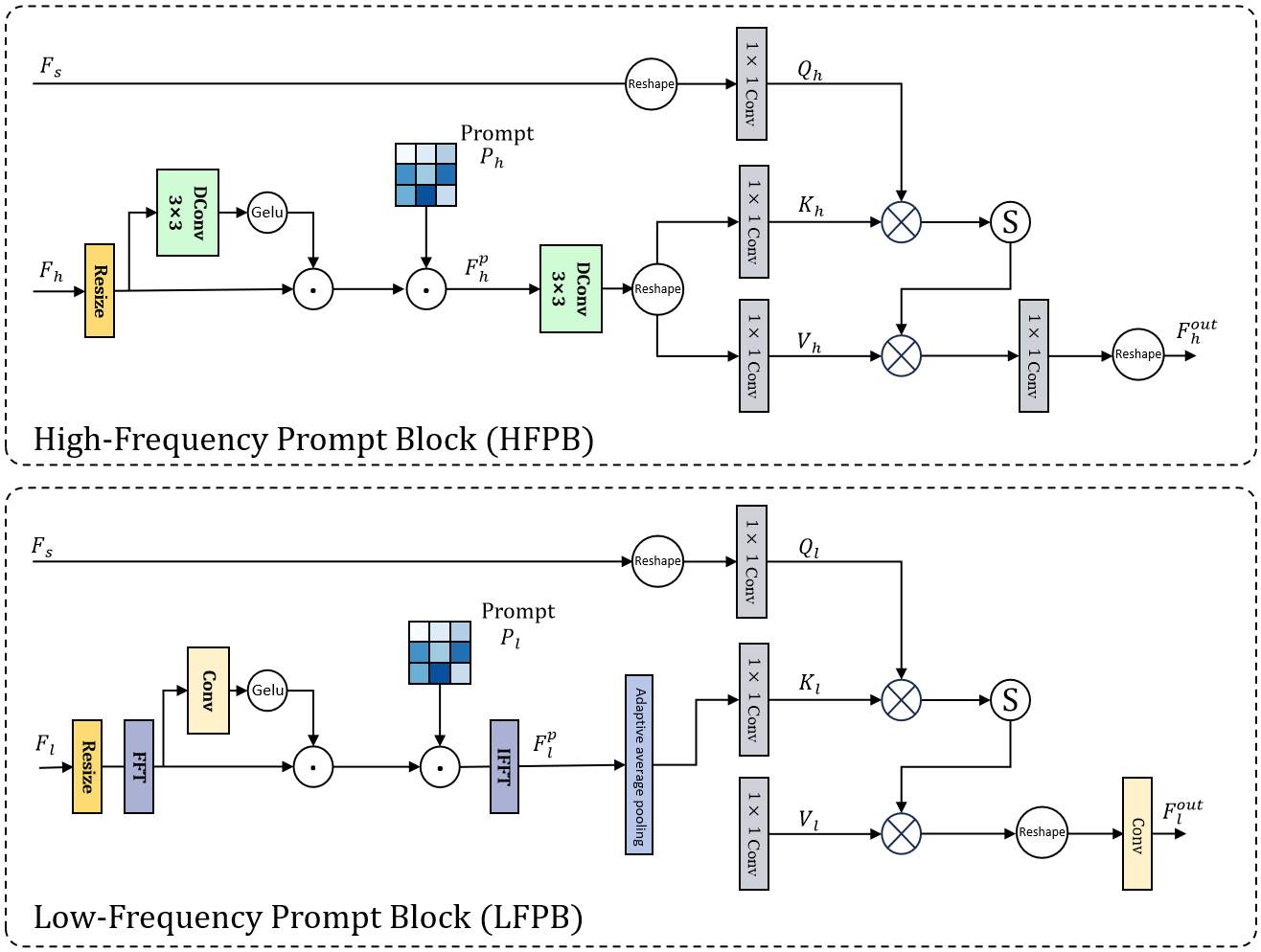}
	\end{center}
	\vspace{-1em}
	\caption{The top and bottom panels illustrate the High-Frequency Prompt Block (HFPB) and Low-Frequency Prompt Block (LFPB), respectively.}
	\vspace{-1em}    
    \label{FreqMoM}
\end{figure}

\noindent\textbf{Unified Frequency Modulation Module (UFMM).}
Although low- and high-frequency components are explicitly decoupled, they are not independent in terms of semantic roles. Both components originate from the same facial geometry and jointly reflect the underlying structural priors of facial landmarks. Motivated by this observation, we propose the UFMM, which modulates low- and high-frequency features in an explicit manner under the All-in-One paradigm.

Given the input feature $\mathbf{F}_s$, UFMM incorporates the high-frequency prior $\mathbf{F}_{\mathrm{h}}$ and the low-frequency prior $\mathbf{F}_{\mathrm{l}}$ into the High-Frequency Prompt Block (HFPB) (Fig.~\ref{FreqMoM}, top) and the Low-Frequency Prompt Block (LFPB) (Fig.~\ref{FreqMoM}, bottom), respectively. These processes can be defined as:
\begin{equation}
\begin{aligned}
\mathbf{Z}_{\mathrm{h}} &= H\!\left(\mathbf{F}_s \mid \mathbf{F}_{\mathrm{h}}\right), \\
\mathbf{Z}_{\mathrm{l}} &= L\!\left(\mathbf{F}_s \mid \mathbf{F}_{\mathrm{l}}\right),
\end{aligned}
\end{equation}
where $(\cdot \mid \cdot)$ denotes that each branch is explicitly modulated by its corresponding frequency prior. Finally, the refined features $\mathbf{Z}_{\mathrm{h}}$ and $\mathbf{Z}_{\mathrm{l}}$ are processed by a concatenation operation along the channel dimension to construct the refined feature $\mathcal{X}_\mathrm{FP}$. Thus, UFMM enables the two branches to model complementary frequency components in a decoupled manner, while both remain governed by the same underlying facial structural prior.

\textbf{(1) High-frequency Prompt Block (HFPB).} Given the input feature $\mathbf{F}_s$ and the high-frequency prior $\mathbf{F}_{\mathrm{h}}$, HFPB aims to selectively enhance fine-grained structural details through frequency-aware prompting and local attention.
Specifically, the high-frequency prior $\mathbf{F}_{\mathrm{h}}$ is first processed by a lightweight depth-wise convolution to aggregate local responses, and is subsequently modulated by a GELU-based gating function to suppress noisy activations while preserving informative high-frequency cues, achieved via element-wise multiplication with the high-frequency prior $\mathbf{F}_{\mathrm{h}}$. The gated feature is then further modulated by a learnable high-frequency prompt $\mathbf{P}_{\mathrm{h}}$, yielding a frequency-enhanced prompt representation. These processes can be formulated as:
\begin{equation}
\mathbf{F}^{\mathrm{p}}_{\mathrm{h}}
=
\big(
\mathbf{F}_{\mathrm{h}} \odot \sigma\!\left(\mathrm{DConv}(\mathbf{F}_{\mathrm{h}})\right)
\big)
\odot \mathbf{P}_{\mathrm{h}},
\end{equation}where $\mathbf{F}^{\mathrm{p}}_{\mathrm{h}}$ denotes the high-frequency prompt feature, serving as a frequency-enhanced representation that highlights geometrically informative local details.

To enhance fine-grained structural modeling, $\mathbf{F}^{\mathrm{p}}_{\mathrm{h}}$ is refined via a cross-attention~\citep{vaswani2017attention} mechanism guided by the input feature $\mathbf{F}_s$. 
This process can be defined as:
\begin{equation}
\mathbf{F}^{\mathrm{out}}_{\mathrm{h}}
=
\mathrm{MHCA}\!\left(\mathbf{F}_s,\mathbf{F}^{\mathrm{p}}_{\mathrm{h}}, \mathbf{F}^{\mathrm{p}}_{\mathrm{h}}\right),
\end{equation}
where $\mathbf{F}_s$ denotes the query, $\mathbf{F}^{\mathrm{p}}_{\mathrm{h}}$ denotes the key and value embeddings. MHCA denotes multi-head cross-attention.  The resulting feature $\mathbf{F}^{\mathrm{out}}_{\mathrm{h}}$ represents the refined high-frequency response, in which fine-grained structural details are selectively emphasized under the guidance of the input feature.

\textbf{(2) Low-frequency Prompt Block (LFPB).} LFPB focuses on modeling global facial geometry and large-scale structural cues.
Given the input feature $\mathbf{F}_s$ and the low-frequency prior $\mathbf{F}_{\mathrm{l}}$, the low-frequency prior is first aligned to the spatial resolution of $\mathbf{F}_s$ and transformed into the frequency domain via a fast Fourier transform (FFT). Then, a lightweight gating operation is applied to regulate the propagation of informative low-frequency responses, followed by modulation with a learnable low-frequency prompt $\mathbf{P}_{\mathrm{l}}$.
The modulated features are subsequently restored to the spatial domain through an inverse FFT, yielding the low-frequency prompt representation:
\begin{equation}
\mathbf{F}^{\mathrm{p}}_{\mathrm{l}}
=
\mathcal{F}^{-1}\!\left(
\big(
\mathcal{F}(\mathbf{F}_{\mathrm{l}})
\odot
\sigma\!\left(\operatorname{Conv}_{1\times1}\big(\mathcal{F}(\mathbf{F}_{\mathrm{l}})\big)\right)
\big)
\odot
\mathbf{P}_{\mathrm{l}}
\right),
\end{equation}
where $\mathcal{F}(\cdot)$ and $\mathcal{F}^{-1}(\cdot)$ denote the FFT and inverse FFT, respectively.

To incorporate structural guidance from the input feature, the low-frequency prompt $\mathbf{F}^{\mathrm{p}}_{\mathrm{l}}$ is further refined via a cross-attention mechanism.
The refinement process is formulated as:
\begin{equation}
\mathbf{F}^{\mathrm{out}}_{\mathrm{l}}
=
\mathrm{MHCA}\!\left(\mathbf{F}_s,\mathbf{F}^{\mathrm{p}}_{\mathrm{l}}, \mathbf{F}^{\mathrm{p}}_{\mathrm{l}}\right),
\end{equation}
where $\mathbf{F}_s$ denotes the query, $\mathbf{F}^{\mathrm{p}}_{\mathrm{l}}$ denotes the key and value embeddings and $\mathbf{F}^{\mathrm{out}}_{\mathrm{l}}$ denotes the refined low-frequency feature that captures coherent global facial structure.

\begin{figure*}[t]
	\begin{center}
		\includegraphics[width=0.75\linewidth]{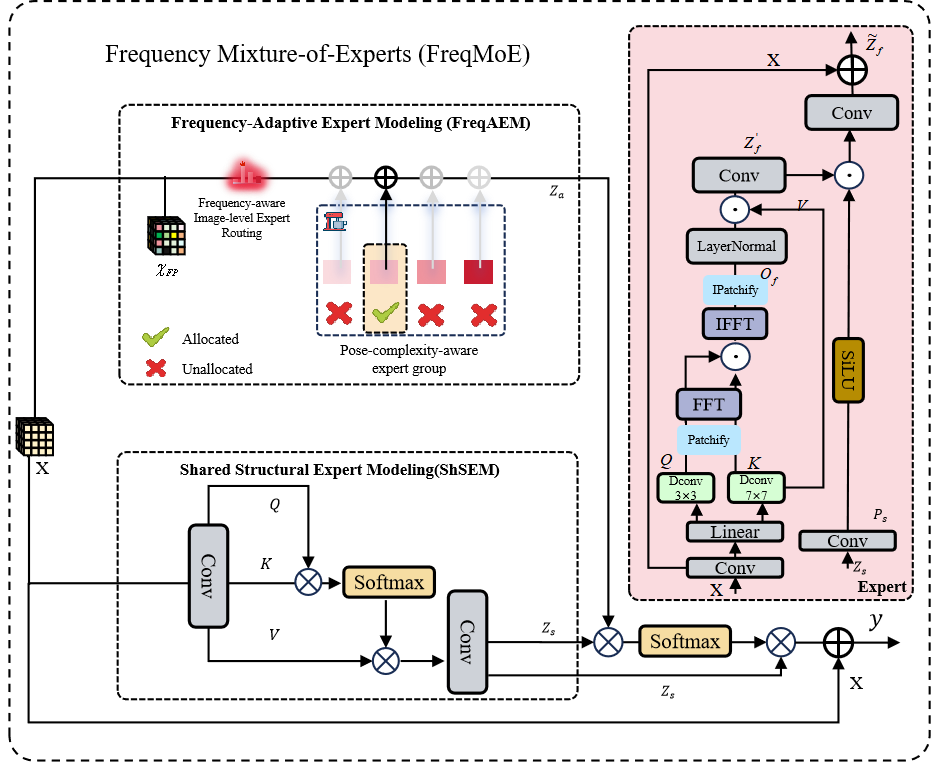}
	\end{center}
    \vspace{-1.5em}
	\caption{The proposed FreqMoE. FreqMoE consists of a Shared Structural Expert Modeling (ShSEM) path and a Frequency-Adaptive Expert Modeling (FreqAEM) path. ShSEM captures unified facial structural representations to preserve common geometric priors, while FreqAEM performs frequency-aware image-level routing and activates pose-complexity-aware experts for modeling sample-specific facial variations. The two paths are finally fused by cross-attention to produce the refined output feature.} 
\label{freqmoe}
\end{figure*}

\subsection{Frequency Mixture-of-Experts (FreqMoE)}
\label{sec:FreqMoE}
FLD is inherently challenged by large variations in facial pose and structural complexity across samples, especially under the All-in-One training paradigm. However, conventional MoE architectures fail to adequately account for such sample-wise heterogeneity, as they typically adopt uniformly configured experts and static capacity allocation. To address this limitation, we propose FreqMoE (Fig.~\ref{freqmoe}), a pose-complexity-aware mixture framework for All-in-One FLD. FreqMoE integrates a unified shared structural expert modeling (ShSEM) path for preserving common facial priors and a frequency-adaptive expert modeling (FreqAEM) path for handling sample-specific pose and frequency variations. Their complementarity enables FreqMoE to unify stable structural prior preservation with complexity-aware frequency adaptation.

\subsubsection{Unified Structural Modeling}
FreqMoE is designed as a unified dual-path modeling framework that integrates ShSEM and FreqAEM. The overall process is expressed as:
\begin{equation}
\begin{aligned}
    \mathbf{z}_s &= \mathrm{ShSEM}(\mathbf{x}), \\
    \mathbf{z}_a &= \mathrm{FreqAEM}(\mathbf{x}, \mathcal{X}_{\mathrm{FP}}, \mathbf{z}_s), \\
    \mathbf{y} &= \mathrm{MHCA}(\mathbf{z}_a, \mathbf{z}_s) + \mathbf{x},
\end{aligned}
\end{equation}
where $\mathrm{ShSEM}(\cdot)$ denotes the shared structural modeling and $\mathrm{FreqAEM}(\cdot)$ denotes the frequency-adaptive expert modeling. This process allows FreqMoE to jointly preserve common facial structural priors and model sample-specific pose-frequency variations.

\subsubsection{Shared Structural Expert Modeling (ShSEM)}
To decouple shared structural representation from sample-specific adaptive modeling, given the input feature $\mathbf{x}$, we first project it through a $1\times1$ convolution and then apply an MHSA mechanism to capture unified facial structural representations. This process can be formulated as:
\begin{equation}
\begin{aligned}
    \mathbf{x}_s &= \phi_s(\mathbf{x}), \\
    \mathbf{z}_s &= \mathrm{MHSA}(\mathbf{x}_s),
\end{aligned}
\end{equation}
where $\phi_s(\cdot)$ denotes the $1\times1$ convolutional projection, $\mathbf{x}_s$ is the projected shared feature, and $\mathbf{z}_s$ denotes the unified structural representation. Through MHSA, $\mathbf{z}_s$ aggregates long-range dependencies among facial components, allowing the ShSEM to encode global facial geometry and consistent structural priors. Therefore, $\mathbf{z}_s$ provides a structure-aware reference for subsequent fusion with the frequency-adaptive expert output.

\subsubsection{Frequency-Adaptive Expert Modeling (FreqAEM)}
To capture sample-specific pose and frequency variations, we introduce the refined frequency-aware feature $\mathcal{X}_{\mathrm{FP}}$ into the adaptive expert modeling. This enables frequency-aware facial geometry modulation by selecting suitable experts in the frequency domain, complementing the unified structural representation $\mathbf{z}_s$.

FreqAEM consists of image-level expert routing and a pose-complexity-aware expert group. The routing mechanism predicts sample-wise expert probabilities from the image-level representation and the refined frequency-aware feature $\mathcal{X}_{\mathrm{FP}}$, enabling adaptive expert allocation according to pose complexity and frequency characteristics. The expert group introduces progressively enlarged frequency receptive fields through different patch sizes, while all experts share the same frequency-modulated architecture, providing multi-scale structural modeling for All-in-One FLD.
The overall process is formulated as:
\begin{equation}
\mathbf{z}_a 
=
\sum_{e=1}^\mathrm{N}
\mathcal{R}_{e}
\left(
\mathbf{x},\mathcal{X}_\mathrm{FP}
\right)
\,
\mathbf{E}_{e}\left( \mathbf{x}, \mathbf{z}_s \right),
\end{equation}
where $\mathrm{N}$ denotes the number of experts, 
$\mathcal{R}_{e}(\mathbf{x},\mathcal{X}_{\mathrm{FP}})$ is the frequency-aware image-level expert routing weight assigned to the $e$-th expert, 
and $\mathbf{E}_{e}\left( \mathbf{x}, \mathbf{z}_s \right)$ denotes the output of the $e$-th frequency-aware expert.


\textbf{(1) Frequency-aware Image-level Expert Routing.}
In FLD, achieving scale-invariant tokenization is particularly challenging, as facial structures must remain consistent across varying resolutions and facial scales. Consequently, conventional token-level routing strategies commonly adopted in prior MoE frameworks~\citep{shazeer2017sparsely,riquelme2021scaling,puigcerver2023sparse} may lead to unstable expert assignment and fragmented structural reasoning. To address this issue, we adopt an image-level routing strategy, where each input image is routed as a whole to a single expert, thereby preserving global facial structure and scale consistency.

Given an input feature $\mathbf{x}$ and the refined frequency-aware feature $\mathcal{X}_{\mathrm{FP}}$, FreqMoE applies a lightweight frequency-conditioned routing function $\mathcal{R}(\cdot)$ to estimate the association between the input representation and each expert. Specifically, the input feature $\mathbf{x}$ is first aggregated into an image-level descriptor by GAP, reshaped into a vector, and then mapped to image-level routing logits through a fully connected layer. Meanwhile, $\mathcal{X}_{\mathrm{FP}}$ is fed into a frequency-aware gate to produce frequency-conditioned routing logits. The two logits are added to produce the final routing logits:
\begin{equation}
\label{eq:routing_logit}
\mathbf{r}=
\mathrm{FC}_{x}
\left(
\mathrm{GAP}(\mathbf{x})
\right)
+
\mathrm{FC}_{f}
\left(
\mathcal{X}_{\mathrm{FP}}
\right)
 ,
\end{equation}
where $\mathbf{r}$ denotes the routing logits, $\mathrm{FC}_{x}(\cdot)$ maps the GAP-based image-level descriptor to expert logits, and $\mathrm{FC}_{f}(\cdot)$ maps the refined frequency-aware feature $\mathcal{X}_{\mathrm{FP}}$ to frequency-conditioned expert logits.

To encourage exploration and mitigate early routing collapse, FreqMoE injects independent Gaussian noise $\boldsymbol{\epsilon}\sim\mathcal{N}(\mathbf{0},\sigma^{2}\mathbf{I})$ into the routing logits. The noisy logits are then normalized by the Softmax operation and sparsified by the Top-$k$ function:
\begin{equation}
\label{eq:routing}
\mathcal{R}(\mathbf{x},\mathcal{X}_{\mathrm{FP}})
=
\operatorname{Top}_{k}
\left(
\operatorname{Softmax}
\left(
\mathbf{r}
+
\boldsymbol{\epsilon}
\right)
\right).
\end{equation}

\textbf{(2) Expert Design.}
Facial pose variations introduce scale-dependent structural complexity, requiring both local landmark details and global facial geometry. Therefore, FreqAEM constructs a pose-complexity-aware expert group with progressively enlarged receptive fields, where experts share the same frequency-modulated structure but use different patch sizes (e.g., smaller patches focus on local landmark details, while larger patches capture broader facial geometry). For each expert, given the feature $\mathbf{x}$ and $\mathbf{z}_s$, we first project them into an expert-specific embedding space:
\begin{equation}
\begin{aligned}
\mathbf{Z}_{f}
&=
\phi_x(\mathbf{x}), \\
\mathbf{P}_{s}
&=
\phi_s(\mathbf{z}_s),
\end{aligned}
\end{equation}
where $\mathbf{Z}_{f}$ denotes the projected input representation, $\mathbf{P}_{s}$ denotes the projected shared representation, and $\phi_x(\cdot)$ and $\phi_s(\cdot)$ are two independent $1\times1$ convolutional projections. 
Then, to capture frequency-structured features, each frequency expert transforms the feature $\mathbf{Z}_{f}$ into query, key, and value representations:
\begin{equation}
\begin{aligned}
\mathbf{Q}
&=
\operatorname{DWConv}_{3\times3}
\left(
\operatorname{Conv}_{1\times1}(\mathbf{Z}_{f})
\right), \\
[\mathbf{K}, \mathbf{V}]
&=
\operatorname{DWConv}_{7\times7}
\left(
\operatorname{Conv}_{1\times1}(\mathbf{Z}_{f})
\right).
\end{aligned}
\end{equation}

Subsequently, $\mathbf{Q}$ and $\mathbf{K}$ are partitioned into local patches and transformed into the frequency domain. Their frequency-domain cross-interaction is computed as:
\begin{equation}
\mathbf{O}_{f}
=
\mathcal{P}_{p_i}^{-1}
\left[
\mathcal{F}^{-1}
\left(
\mathcal{F}\left(\mathcal{P}_{p_i}(\mathbf{Q})\right)
\odot
\mathcal{F}\left(\mathcal{P}_{p_i}(\mathbf{K})\right)
\right)
\right],
\end{equation}
where $\mathcal{P}_{p_i}(\cdot)$ denotes patch partition with patch size 
$p_i \in \{2^{i+2}\}_{i=0}^{N-1}$ for the $i$-th frequency expert, and $\mathcal{P}_{p_i}^{-1}(\cdot)$ denotes the inverse patch rearrangement, $\mathcal{F}(\cdot)$ and $\mathcal{F}^{-1}(\cdot)$ denote FFT and inverse FFT, respectively, and $\odot$ denotes element-wise multiplication in the frequency domain. After obtaining the frequency-domain interaction response $\mathbf{O}_{f}$, we further normalize it and modulate it with the value representation $\mathbf{V}$ to inject spatial content information:
\begin{equation}
\mathbf{Z}'_{f}
=
\phi_o
\left(
\mathrm{LN}(\mathbf{O}_{f})
\odot
\mathbf{V}
\right),
\end{equation}
where $\phi_o(\cdot)$ denotes the output $1\times1$ convolution, and $\mathbf{Z}'_{f}$ is the frequency-enhanced representation, which is then reweighted by the shared modulation signal $\mathbf{P}_s$ and combined with the input through a residual connection. This process is defined as:
\begin{equation}
\tilde{\mathbf{Z}}_{f}
=
\rho
\left(
\mathbf{Z}'_{f}
\odot
\sigma(\mathbf{P}_{s})
\right)
+
\mathbf{x},
\end{equation}
where $\sigma(\cdot)$ denotes the SiLU activation, $\rho(\cdot)$ denotes the output projection, and $\odot$ denotes element-wise multiplication.

The above operation integrates frequency-refined cues with expert-specific representations, enabling fine-grained FLD under the All-in-One paradigm across diverse facial variations and imaging conditions.

\subsection{ Frequency-Consistent Routing (FreqCR) Loss}
\label{sec:Loss}
In MoE architectures, experts with heterogeneous computational capacities are susceptible to imbalanced utilization, routing instability, and degenerate expert collapse. These issues are further exacerbated under the All-in-One paradigm, where a single model is required to generalize across diverse facial conditions and varying levels of structural and frequency complexity, leading to pronounced disparities in expert activation and routing behavior. To address these issues, we introduce a FreqCR loss, which explicitly regularizes expert utilization to be both balanced across experts and consistent with their computational capacities. The FreqCR loss is defined as:
\begin{equation}
\mathcal{L}_{\text{FreqCR}} =
\frac{1}{2}\mathcal{L}_{\text{imp}} +
\frac{1}{2}\mathcal{L}_{\text{load}},
\end{equation}
where $\mathcal{L}_{\text{imp}}$ enforces balanced expert importance under a complexity-aware bias, and $\mathcal{L}_{\text{load}}$ regularizes the effective expert assignment frequency during routing.

\textbf{(1) Complexity-aware importance loss.}
To incorporate expert computational capacity into routing, each expert is associated with a pre-computed complexity prior $\boldsymbol{\theta}$, where each element $\theta_e \in \boldsymbol{\theta}$ denotes the number of learnable parameters of the $e$-th expert and is fixed during training. 
A normalized complexity bias is then defined as $\boldsymbol{\beta} = \boldsymbol{\theta} / \theta_{\max}$.
For a mini-batch $\mathcal{B}$, the complexity-aware importance loss is formulated as:
\begin{equation}
\mathcal{L}_{\mathrm{imp}}
=
\mathrm{COV}
\left(
\left(
\sum_{x \in \mathcal{B}} \mathrm{Softmax}(\mathcal{R}(x))
\right)
\odot \boldsymbol{\beta}
\right)^{2},
\end{equation}
where $\mathrm{Softmax}(\mathcal{R}(x))$ yields the normalized importance weights over experts, $\boldsymbol{\beta}\in\mathbb{R}^{n}$ is the normalized complexity bias derived from the expert parameter counts, $\odot$ denotes element-wise multiplication, and $\mathrm{COV}(\cdot)$~\citep{riquelme2021scaling} denotes the coefficient of variation computed across experts.

\textbf{(2) Load balancing loss.}
To mitigate routing imbalance, we estimate the expected expert load using a probabilistic approximation under noisy Top-$k$ routing.
Given routing logits $\mathcal{R}(x)\in\mathbb{R}^{n}$, we inject Gaussian noise
$\boldsymbol{\epsilon}\sim\mathcal{N}(\mathbf{0},\boldsymbol{\sigma}^2)$
and obtain noisy logits, which can be formulated as:
\begin{equation}
  \tilde{\mathcal{R}}(x)=\mathcal{R}(x)+\boldsymbol{\epsilon}.  
\end{equation}

Let $\tau(x)$ denote the selection threshold defined as the $k$-th largest value in
$\tilde{\mathcal{R}}(x)$.
The expected expert load over a mini-batch $\mathcal{B}$ and the corresponding load balancing loss are defined as:
\begin{equation}
\mathcal{L}_{\mathrm{load}}
=
\mathrm{COV}\!\left(
\frac{1}{|\mathcal{B}|}
\sum_{x\in\mathcal{B}}
\Big(
\mathbf{1}
-
\Phi\!\left(
\frac{\tau(x)\mathbf{1}-\mathcal{R}(x)}{\boldsymbol{\sigma}}
\right)
\Big)
\right)^{2},
\end{equation}
where $\mathbf{1}\in\mathbb{R}^{n}$ is an all-ones vector used for threshold broadcasting,
$\boldsymbol{\sigma}$ denotes the per-expert noise standard deviation,
$\Phi(\cdot)$ is the standard normal cumulative distribution function (CDF) used to approximate expert selection probability under noisy Top-$k$ routing,
and $\mathrm{COV}(\cdot)$ measures the coefficient of variation across experts.

Finally, the overall training objective is defined by integrating a  MSE loss~\citep{park2020complementary} into the optimization, which is formulated as:
\begin{equation}
\mathcal{L}_{\text{total}}
=
\lambda_1 \mathcal{L}_{\text{MSE}}
+
\lambda_2 \mathcal{L}_{\text{FreqCR}},
\end{equation}
where $\lambda_1$ and $\lambda_2$ control the corresponding weights of the individual loss terms, respectively.

\section{EXPERIMENTS}\label{EXPERIMENTS}
We evaluate the proposed method on four widely adopted FLD benchmarks, including 300W~\citep{300w}, COFW~\citep{cofw}, WFLW~\citep{wflw}, and AFLW~\citep{aflw}. Comparisons are conducted against state-of-the-art FLD approaches, where $\circ$ and $\diamond$ denote heatmap regression and coordinate regression methods, respectively. We further present comprehensive ablation studies and self-evaluation analyses to systematically examine the effectiveness of the proposed FreqFLD framework.

\subsection{Datasets and Implementation Details}

\paragraph{\textbf{300W (68 landmarks)}~\citep{300w}} The 300W dataset is a widely adopted benchmark for facial landmark detection. It comprises 3,148 training images and 689 testing images, with each face annotated using 68 landmarks. The test set is further divided into two subsets: a Common subset and a Challenging subset. The Common subset is composed of 224 images drawn from the LFPW~\citep{belhumeur2013localizing} test set and 330 images from the HELEN~\citep{le2012interactive} test set, while the Challenging subset contains 135 images from the IBUG~\citep{300w} dataset, which exhibit significant pose variations and occlusions.

\paragraph{\textbf{WFLW (98 landmarks)}~\citep{wflw}}
WFLW consists of 7,500 training images and 2,500 testing images, each annotated with 98 facial landmarks. The test set is further categorized into multiple subsets according to diverse facial variations, including pose, expression, illumination, and occlusion. This dataset provides a comprehensive benchmark for evaluating the robustness of FLD methods.

\paragraph{\textbf{COFW (29 landmarks)}~\citep{cofw}}
COFW is a benchmark dataset designed for evaluating facial landmark detection in the presence of heavy occlusions. The dataset comprises 1,345 facial images annotated with 29 landmarks. Among them, 845 images are used for training and the remaining 500 images are used for testing.

\paragraph{\textbf{AFLW (19 landmarks)}~\citep{aflw}}
The AFLW dataset contains 24,386 face images with large pose variations.
Following standard evaluation protocols, we use 19 facial landmarks for evaluation.

\begin{table}[t]
    \centering
    \scriptsize
    \caption{Comparison of state-of-the-art methods on the 300W dataset, with NME normalized by inter-ocular distance. (\% omitted)}
   \setlength{\tabcolsep}{11pt}
    \renewcommand\arraystretch{1}
    \begin{tabular}{p{6cm}|ccc}
     \toprule[1pt]
        Method  & Common  & Challenging  & Full  \\
        \midrule
        \makebox[3.8cm][l]{$\circ$\, HGs (ECCV16) \citep{yang2017stacked}} & 3.72 & 7.23 & 4.41 \\
        \makebox[3.8cm][l]{$\circ$\, MDM (CVPR16) \citep{trigeorgis2016mnemonic}} & 4.36 & 7.56 & 4.99 \\
        \makebox[3.8cm][l]{$\circ$\, FAN (ICCV17) \citep{bulat2017far}} & 3.08 & 5.52 & 3.56 \\
        \makebox[3.8cm][l]{$\circ$\, LAB (CVPR18) \citep{wflw}} & 2.98 & 5.19 & 3.49 \\
        \makebox[3.8cm][l]{$\circ$\, Wing (CVPR18) \citep{feng2018wing}} & 2.93 & 5.23 & 3.38 \\
        \makebox[3.8cm][l]{$\circ$\, ODN (CVPR19) \citep{zhu2019robust}} & 3.56 & 6.67 & 4.17 \\
        \makebox[3.8cm][l]{$\diamond$\, AWing (ICCV19) \citep{wang2019adaptive}} & 2.72 & 4.52 & 3.07 \\
         \makebox[3.8cm][l]{$\diamond$\, LUVLi (CVPR20) \citep{kumar2020luvli}} & 2.76 & 5.16 & 3.23 \\
        \makebox[3.8cm][l]{$\diamond$\, SAAT (ICCV21) \citep{zhu2021improving}} & 2.82 & 5.03 & 3.25 \\
        \makebox[3.8cm][l]{$\diamond$\, SDFL (TIP21) \citep{lin2021structure}} & 2.88 & 4.93 & 3.28 \\
        \makebox[3.8cm][l]{$\diamond$\, SLPT (CVPR22) \citep{xia2022sparse}} & 2.75 & 4.90 & 3.17 \\
        \makebox[3.8cm][l]{$\diamond$\, GlomFace (CVPR22) \citep{zhu2022occlusion}} & 2.72 & 4.79 & 3.13 \\
        \makebox[3.8cm][l]{$\diamond$\, PicassoNet (TNNLS23) \citep{wen2022picassonet}} & 3.03 & 5.81 & 3.58 \\
        \makebox[3.8cm][l]{$\diamond$\, GFL (CVPR24) \citep{liang2024generalizable}} & 2.79 & 4.91 & 3.20 \\
        \hline
        \rowcolor{my_color}\makebox[3.8cm][l]{$\circ$\, \textbf{FreqFLD (ours)}} & 
        \textbf{2.72} & \textbf{4.52} & \textbf{3.08} \\
     \bottomrule[1pt]
    \end{tabular}
    \label{300wtable}
\end{table}

\begin{table}[t]
  \centering
  \scriptsize
  \caption{Comparison of state-of-the-art methods on the COFW dataset, with NME normalized by inter-pupil distance (\% omitted).}
  \setlength{\tabcolsep}{8pt}
  \renewcommand\arraystretch{0.95}
  \begin{tabularx}{\columnwidth}{X c c}
      \toprule
      Method & $\rm NME_{ip}$ & FR \\
      \midrule
      $\circ$\, Wing (CVPR18)~\citep{feng2018wing} & 5.44 & 3.75 \\
      $\circ$\, DCFE (ECCV18)~\citep{valle2018deeply} & 5.27 & 0.35 \\
      $\circ$\, AWing (ICCV19)~\citep{wang2019adaptive} & 4.94 & 0.99 \\
      $\circ$\, ODN (CVPR19)~\citep{zhu2019robust} & 5.30 & -- \\
      $\circ$\, MHHN (TIP20)~\citep{wan2020robust} & 4.95 & 1.78 \\
      $\diamond$\, ADNet (ICCV21)~\citep{huang2021adnet} & 4.68 & 0.59 \\
      $\diamond$\, MMDN (TNNLS22)~\citep{wan2021robust} & 5.01 & 1.78 \\
      $\diamond$\, SLPT (CVPR22)~\citep{xia2022sparse} & 4.79 & 1.18 \\
      $\diamond$\, DSLPT-R50 (TPAMI23)~\citep{xia2023robust} & 4.81 & 1.18 \\
      $\diamond$\, CIT-v2 (IJCV24)~\citep{li2024cascaded} & 5.81 & 3.55 \\
      \midrule
      \rowcolor{my_color}
      $\circ$\, \textbf{FreqFLD (ours)} & \textbf{4.80} & \textbf{0.39} \\
      \bottomrule
  \end{tabularx}
  \label{cofwtable}
\end{table}

\begin{table*}[t]
  \centering
  \scriptsize
  \caption{Comparison of state-of-the-art methods on the WFLW dataset, with NME normalized by inter-ocular distance (\% omitted).}
    \setlength{\tabcolsep}{1pt}
  \resizebox{\textwidth}{!}{%
  \begin{tabular}{p{5.6cm}|ccccccc}
      \toprule[1pt]
      {\normalsize Method}
      & {\normalsize Testset}
      & {\normalsize\makecell{Pose\\Subset}}
      & {\normalsize\makecell{Expression\\Subset}}
      & {\normalsize\makecell{Illumination\\Subset}}
      & {\normalsize\makecell{Make-Up\\Subset}}
      & {\normalsize\makecell{Occlusion\\Subset}}
      & {\normalsize\makecell{Blur\\Subset}} \\
      \midrule
        \makebox[5.0cm][l]{$\circ$\, ESR \citep{cao2014face}} 
        & 11.13 & 25.88 & 11.47 & 10.49 & 11.05 & 13.75 & 12.20 \\
        \makebox[5.0cm][l]{$\circ$\, SDM \citep{xiong2013supervised}} 
        & 10.29 & 24.10 & 11.45 & 9.32 & 9.38 & 13.03 & 11.28 \\
        \makebox[5.0cm][l]{$\circ$\, CFSS \citep{zhu2015face}} 
        & 9.07 & 21.36 & 10.09 & 8.30 & 8.74 & 11.76 & 9.96 \\
        \makebox[5.0cm][l]{$\circ$\, LAB (CVPR18) \citep{wflw}} 
        & 5.27 & 10.24 & 5.51 & 5.23 & 5.15 & 6.79 & 6.32 \\
        \makebox[5.0cm][l]{$\circ$\, Wing (CVPR18) \citep{feng2018wing}} 
        & 5.11 & 8.75 & 5.36 & 4.93 & 5.41 & 6.37 & 5.81 \\
        \makebox[5.0cm][l]{$\circ$\, DeCaFA (ICCV19) \citep{dapogny2019decafa}} 
        & 4.62 & 8.11 & 4.65 & 4.41 & 4.63 & 5.74 & 5.38 \\
        \makebox[5.0cm][l]{$\diamond$\, HRNet (TPAMI20) \citep{wang2020deep}} 
        & 4.60 & 7.86 & 4.78 & 4.57 & 4.26 & 5.42 & 5.36 \\
        \makebox[5.0cm][l]{$\diamond$\, MHHN (TIP20) \citep{wan2020robust}} 
        & 4.77 & 9.31 & 4.79 & 4.72 & 4.59 & 6.17 & 5.82 \\
        \makebox[5.0cm][l]{$\diamond$\, MMDN (TNNLS21) \citep{wan2021robust}} 
        & 4.87 & 7.71 & 4.79 & 4.61 & 4.72 & 6.17 & 5.72 \\
        \makebox[5.0cm][l]{$\diamond$\, GlomFace (CVPR22) \citep{zhu2022occlusion}} 
        & 4.81 & 8.71 & -- & -- & -- & 5.14 & -- \\
        \makebox[5.0cm][l]{$\diamond$\, EfficientFan (TNNLS23) \citep{gao2021facial}} 
        & 4.54 & 8.20 & 4.87 & 4.39 & 4.54 & 5.42 & 5.04 \\
        \makebox[5.0cm][l]{$\diamond$\, PicassoNet (TNNLS23) \citep{wen2022picassonet}} 
        & 4.82 & 8.61 & 5.14 & 4.73 & 4.68 & 5.91 & 5.56 \\
      \hline
        \rowcolor{my_color}
        \makebox[5.0cm][l]{$\circ$\, \textbf{FreqFLD (ours)}} 
        & \textbf{4.50} & \textbf{7.54} & \textbf{4.59} & \textbf{4.55} 
        & \textbf{4.46} & \textbf{5.53} & \textbf{5.15} \\
      \bottomrule[1pt]
  \end{tabular}}
  \label{tabwflw}
\end{table*}

\begin{table}[t]
  \centering
  \scriptsize
  \caption{Comparison of state-of-the-art methods on the AFLW dataset, with NME normalized by face size. (\% omitted)}
  \renewcommand\arraystretch{1}
    \setlength{\tabcolsep}{8pt}
  \begin{tabularx}{\columnwidth}{X c}
      \toprule[1pt]
      Method & \bf $\mathrm{NME_{box}}$ \\
      \midrule
      $\circ$\, DAC-CSR (CVPR17)~\cite{feng2017dynamic} & 2.27 \\
      $\circ$\, SAN (CVPR18)~\cite{dong2018style} & 1.91 \\
      $\circ$\, LAB (CVPR18)~\cite{wflw} & 1.85 \\
      $\circ$\, LLL (ICCV19)~\cite{robinson2019laplace} & 1.97 \\
      $\circ$\, LUVLi (CVPR20)~\cite{kumar2020luvli} & 1.39 \\
      $\circ$\, HRNet (TPAMI20)~\cite{wang2020deep} & 1.57 \\
      $\circ$\, MHHN (TIP20)~\cite{wan2020robust} & 1.38 \\
      $\diamond$\, PIPNet (IJCV21)~\cite{jin2021pixel} & 1.42 \\
      $\diamond$\, PicassoNet (TNNLS23)~\cite{wen2022picassonet} & 1.59 \\
      $\diamond$\, Protoformer (TMM26)~\cite{hu2026proto} & 1.47 \\
      \midrule
      \rowcolor{my_color} $\circ$\, \textbf{FreqFLD (ours)} & \textbf{1.69} \\
      \bottomrule[1pt]
  \end{tabularx}
  \label{aflwtable}
\end{table}

\paragraph{\textbf{Evaluation Metrics}} FreqFLD adopts the Normalized Mean Error (NME) as the primary evaluation metric for FLD. 
Following standard practice, different normalization factors are used for different datasets: the inter-ocular distance is adopted for 300W and WFLW, the face bounding box size is used for AFLW, and the inter-pupil distance is employed for COFW. 
In addition, we also report the Failure Rate (FR)  for the COFW dataset.

\paragraph{\textbf{Implementation Details}} All input images are resized to $256 \times 256 \times 3$ in our experiments. The loss weights $\lambda_1$ and $\lambda_2$ are set to 1 and $1 \times 10^{-4}$, respectively. During training, data augmentation is applied to enhance model robustness, including random rotations within $\pm 30^\circ$ and horizontal flipping with a probability of 0.5. The proposed FreqFLD is implemented in PyTorch and trained on an NVIDIA RTX 4090 GPU for 100,000 iterations with a batch size of 8. We employ the AdamW optimizer with an initial learning rate of $1 \times 10^{-4}$. 

\paragraph{\textbf{All-in-One Training Paradigm}}
Beyond the conventional paradigm, we further extend our evaluation to an All-in-One training setting, where multiple datasets are jointly used to train a single unified model. In this setting, the training data from 300W, AFLW, COFW, and WFLW are combined, and each dataset is expanded to contain an equal number (20,000) of training samples, resulting in a balanced training set of 80,000 images in total. The unified model is then separately evaluated on each test set following the standard protocol of the corresponding dataset.

\subsection{Quantitative analysis}
In this section, we quantitatively evaluate the proposed FreqFLD on diverse and challenging scenarios under the All-in-One training paradigm.

\paragraph{\textbf{Evaluations under Normal Circumstances}}
Under normal circumstances, we conduct comparative evaluations on the 300W and AFLW benchmarks, which mainly contain favorable facial images. FreqFLD achieves an $\rm NME_{io}$ of 2.72 on the 300W Common subset and 3.08 on the 300W Full set (Table~\ref{300wtable}), achieving comparable FLD performance across heatmap-based and coordinate-regression-based methods. Moreover, on the WFLW benchmark, FreqFLD achieves competitive performance across multiple subsets (Table~\ref{tabwflw}). These results can be attributed to the proposed FreqMoM, which explicitly disentangles and enhances frequency-aware facial representations, the FreqMoE that adaptively models heterogeneous landmark patterns, and the FreqCR loss that regularizes expert assignment and stabilizes frequency-consistent learning.

\paragraph{\textbf{Evaluation of Robustness against Occlusion}}
To evaluate robustness under occluded scenarios, we conduct experiments on the COFW dataset, the 300W challenging subset, and the WFLW Occlusion subset. On the COFW benchmark (Table~\ref{cofwtable}), FreqFLD achieves an $\rm NME_{ip}$ of 4.80 with a failure rate of 0.39, outperforming most recent FLD methods~\citep{wang2019adaptive,zhu2019robust,xia2023robust,li2024cascaded}. Moreover, FreqFLD attains an $\rm NME_{io}$ of 5.53 (Table~\ref{tabwflw}), demonstrating consistent improvements over existing approaches~\citep{wflw,feng2018wing,dapogny2019decafa,wan2021robust,yu2025helpnet}. Similar performance gains are observed on the 300W challenging subset, where FreqFLD maintains stable accuracy and competitive robustness compared with prior methods~\citep{lin2021structure,xia2022sparse,zhu2022occlusion,wen2022picassonet,liang2024generalizable,yu2025helpnet}. These results are mainly attributed to the proposed FreqMoE. By adaptively activating frequency-aware experts, FreqMoE preserves complementary low-frequency structural cues and high-frequency landmark-sensitive details, enabling stable FLD performance.

\paragraph{\textbf{Evaluation of Robustness against Large Poses} }
Faces with large pose variations or extreme expressions introduce complex geometric distortions. To assess the performance of FreqFLD under these challenging conditions, we conduct evaluations on WFLW-Pose subset, the AFLW full set, and the 300W challenge subset. As shown in Table~\ref{300wtable}, FreqFLD achieves an $\rm NME_{io}$ of 4.52 on the 300W challenging subset, outperforming several SOTA methods~\citep{zhu2021improving,lin2021structure,xia2022sparse,zhu2022occlusion,wen2022picassonet,liang2024generalizable,yu2025helpnet}. Moreover, FreqFLD achieves comparable NME on the AFLW full set (Table~\ref{aflwtable}) and superior performance on the pose and expression subsets of WFLW (Table~\ref{tabwflw}) compared with existing approaches~\citep{feng2018wing,dapogny2019decafa,wang2020deep,gao2021facial,wen2022picassonet}, further confirming its robustness to extreme facial variations. These experimental results indicate that the performance gains of FreqFLD arise from two complementary aspects. By explicitly modeling facial structural cues, FreqFLD exhibits enhanced robustness to large pose and expression variations. Meanwhile, the proposed FreqMoE estimates the pose complexity of input samples and adaptively routes them to experts with appropriate capacity, thereby contributing to robust facial landmark detection performance.


\paragraph{\textbf{Evaluation of Robustness against Blur}}
Faces captured under low-light conditions or affected by blur suffer from substantial loss of visual details, which poses considerable challenges for accurate FLD. To evaluate the effectiveness of our proposed FreqFLD, we conduct experiments on the 300W challenging subset and WFLW-blur subset. As reported in Table~\ref{300wtable}, FreqFLD achieves an $\rm NME_{io}$ of 4.52 on the 300W challenging subset, outperforming competing approaches. Furthermore, as shown in Table~\ref{tabwflw}, FreqFLD attains $\rm NME_{io}$ scores of 4.55 and 5.15 on the illumination and blur subsets of WFLW, respectively, surpassing several representative methods~\citep{feng2018wing,wang2020deep,wan2021robust,wen2022picassonet}. These results demonstrate that FreqFLD consistently maintains high FLD accuracy even under poor image quality. This robustness can be largely attributed to the proposed FreqMoM, which effectively preserves and enhances facial structural cues, enabling reliable landmark prediction despite severe blur and illumination degradation.

\subsection{Ablation Study}
The ablation studies are conducted to analyze the impact of FreqMoE, FreqMoM and FreqCR loss, together with the effect of multi-dataset joint training. The detailed results are presented as follows.

\begin{table}[t]
  \centering
  \scriptsize
  \caption{Influence of FreqMoE, FreqMoM and FreqCR on the 300W challenging subset.}
  \renewcommand\arraystretch{1}
  \setlength{\tabcolsep}{6.5pt}
  \begin{tabular}{c c c c c c}
      \toprule[1pt]
      Method & TB & FreqMoE & FreqMoM & FreqCR& \bf $\mathrm{NME_{io}}$ \\
      \midrule
      Trans (baseline)
      & \ding{51} &  &  && 4.71 \\

      Trans + FreqMoE
      & \ding{51} & \ding{51} &  && 4.60 \\

      Trans + FreqMoM
      & \ding{51} &  & \ding{51} && 4.62 \\

      \rowcolor{my_color}
      Trans + FreqMoE + FreqMoM + FreqCR
      & \ding{51} & \ding{51} & \ding{51}& \ding{51} & \textbf{4.52} \\
      \bottomrule[1pt]
  \end{tabular}
  \label{tab_ceab_fpm}
\end{table}

\paragraph{\textbf{Influence of FreqMoM, FreqMoE and FreqCR loss}}
As shown in Table~\ref{tab_ceab_fpm}, we conduct an ablation study by progressively incorporating FreqMoE, FreqMoM, and FreqCR into the baseline Trans on the 300W challenging subset. The baseline Trans achieves an $\mathrm{NME_{io}}$ of 4.71. By introducing FreqMoE, the NME is reduced to 4.60. Similarly, adding FreqMoM yields an $\mathrm{NME_{io}}$ of 4.62, improving the baseline by 0.09. When FreqMoE and FreqMoM are jointly employed and further equipped with the FreqCR loss, the full model achieves the best performance with an $\mathrm{NME_{io}}$ of 4.52. This result surpasses Trans + FreqMoE and Trans + FreqMoM by margins of 0.08 and 0.10, respectively, and leads to an overall improvement of 0.19 compared with the baseline. 

These results can be attributed to: 1) The introduced FreqMoE effectively strengthens the model’s capability to adapt to faces of differing complexities. 2) By integrating high- and low-frequency facial cues, the FreqMoM produces refined feature representations, thereby improving the model's adaptability to diverse facial geometries. By integrating Trans, FreqMoM and FreqMoE, the model achieves high-precision FLD across different datasets.

\begin{table}[t]
  \centering
  \scriptsize
  \caption{Influence of multi-dataset training on the 300W challenging subset.}
  \setlength{\tabcolsep}{22pt}
  \renewcommand\arraystretch{1.0}
  \begin{tabular}{c c c c c}
    \toprule
    300W & AFLW & WFLW & COFW & $\mathrm{NME_{io}}$ \\
    \midrule
    \ding{51} &  &  &  & 4.97 \\
    \ding{51} & \ding{51} &  &  & 4.75 \\
    \ding{51} & \ding{51} & \ding{51} &  & 4.64 \\
    \rowcolor{my_color}
    \ding{51} & \ding{51} & \ding{51} & \ding{51} & \textbf{4.52} \\
    \bottomrule
  \end{tabular}
  \label{tab:multi_dataset}
\end{table}

\paragraph{\textbf{Influence of Multi-dataset Training}}
We evaluate the effect of multi-dataset training by conducting experiments with different combinations of datasets. Using 300W as the baseline, AFLW, WFLW, and COFW are gradually incorporated into the training set. As shown in Table \ref{tab:multi_dataset}, this progressive inclusion leads to performance improvements of 0.22, 0.33, and 0.45, respectively, indicating that leveraging multiple datasets consistently enhances model performance.

\paragraph{\textbf{Leave-One-Dataset-Out Cross-Dataset Evaluation}}
To verify FreqFLD learns generalizable representations across datasets, we conduct a leave-one-dataset-out evaluation. In each setting, one dataset is excluded from training and used only for testing, while the remaining three datasets are used for model optimization. As shown in Tab.~\ref{tab:lodo_generalization}, FreqFLD maintains stable performance across all held-out datasets, achieving $\mathrm{{NME}_{io}}$ of 4.74 on 300W and 4.85 on WFLW, $\mathrm{{NME}_{ip}}$ of 5.13 on COFW, and $\mathrm{{NME}_{box}}$ of 1.88 on AFLW. These results indicate that FreqFLD can effectively transfer learned frequency-aware facial priors to unseen annotation distributions and facial appearance statistics, demonstrating the cross-dataset generalization ability of FreqFLD.
\begin{table}[t]
  \centering
  \scriptsize
  \caption{Leave-one-dataset-out cross-dataset generalization evaluation. (\% omitted)}
  \renewcommand\arraystretch{1}
  \setlength{\tabcolsep}{4pt}
  \begin{tabular}{l p{0.36\columnwidth} c c}
      \toprule[1pt]
      Setting & Training Datasets & Test Dataset & \bf NME \\
      \midrule
      LODO-300W 
      & WFLW + COFW + AFLW 
      & 300W 
      & 4.74  \\
      
      LODO-WFLW 
      & 300W + COFW + AFLW 
      & WFLW 
      & 4.85  \\
      
      LODO-COFW 
      & 300W + WFLW + AFLW 
      & COFW 
      & 5.13  \\
      
      LODO-AFLW 
      & 300W + WFLW + COFW 
      & AFLW 
      & 1.88 \\
      \bottomrule[1pt]
  \end{tabular}
  \label{tab:lodo_generalization}
\end{table}

\begin{table}[t]
  \centering
  \scriptsize
  \caption{Influence of different numbers of experts and TopK values on the 300W challenging subset (\% omitted).}
  \setlength{\tabcolsep}{38pt} 
  \begin{tabular}{c c | c}
      \toprule[1pt]
      Number of experts & TopK & $\mathrm{NME_{io}}$ \\
      \midrule
      16 & 1 & 4.95 \\
      12 & 1 & 4.82 \\
       8 & 1 & 4.65 \\
       6 & 3 & 4.75 \\
       4 & 2 & 4.60 \\
       2 & 1 & 4.87 \\
      \rowcolor{my_color}
       4 & 1 & \textbf{4.52} \\
      \bottomrule[1pt]
  \end{tabular}
  \label{expert_num}
\end{table}

\begin{table}[t]
  \centering
  \scriptsize
  \caption{The effect of different frequency prompts on the 300W challenging subset (\% omitted).}
  \setlength{\tabcolsep}{38pt} 
  \begin{tabular}{c | c}
      \toprule[1pt]
      Method & $\mathrm{NME_{io}}$ \\
      \midrule
      Trans (baseline) + FreqMoE & 4.60 \\
      Trans + FreqMoE + HFPB & 4.56 \\
      Trans + FreqMoE + LFPB & 4.58 \\
      \rowcolor{my_color}
      Trans + FreqMoE + FreqMoM (HFPB + LFPB) & \textbf{4.52} \\
      \bottomrule[1pt]
  \end{tabular}
  \label{tab_sensi}
\end{table}

\begin{figure}[t]
\begin{center}
	\includegraphics[width=0.7\linewidth]{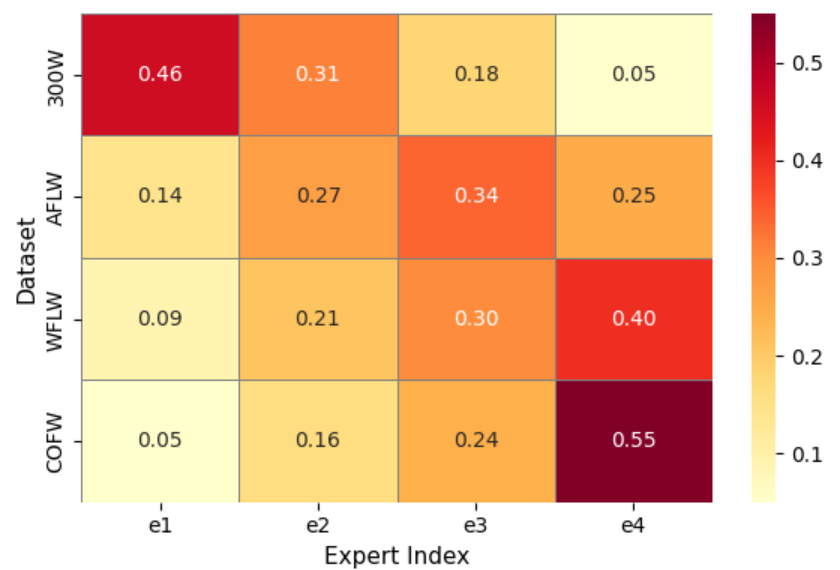}
	\end{center}
	\vspace{-1.5em}
	\caption{Visualization of the complexity experts across different facial landmark datasets.}
	\label{experts}
	\vspace{-1em}    
\end{figure}

\begin{figure*}[t]
\begin{center}
	\includegraphics[width=1\linewidth]{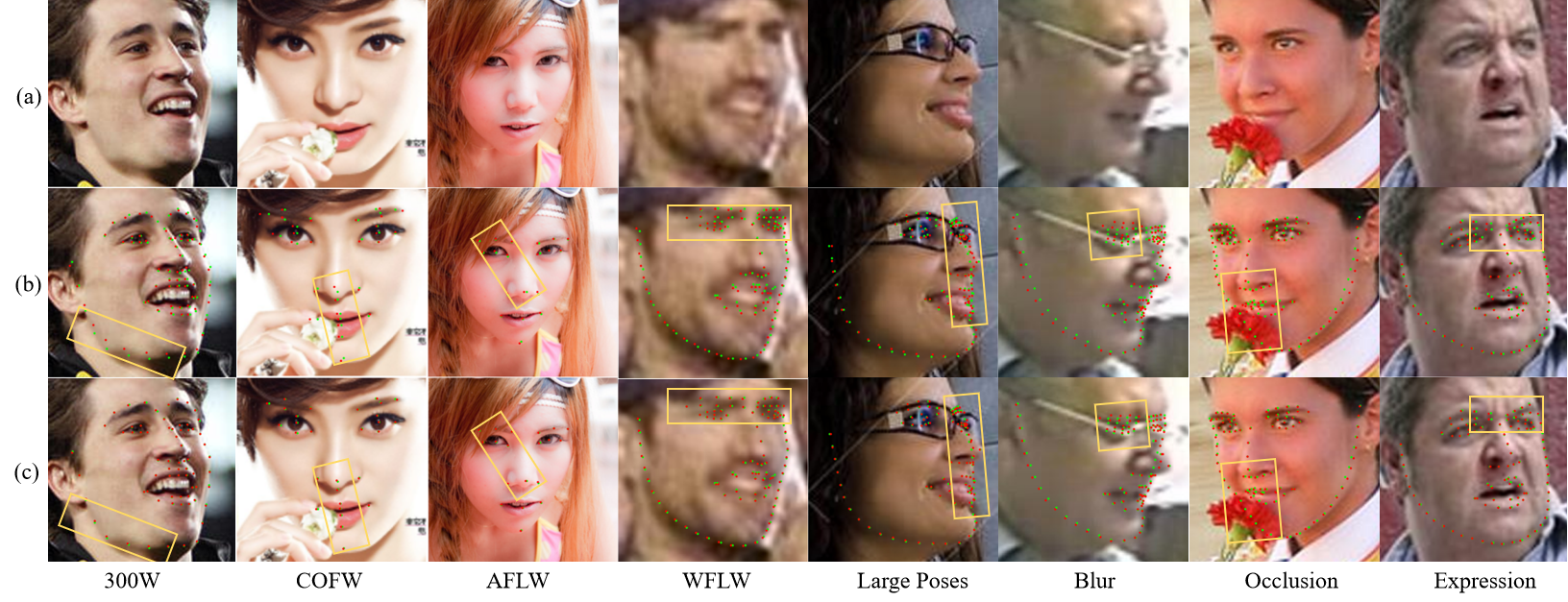}
	\end{center}
	\vspace{-2em}
	\caption{(a) The original input images, (b) the prediction results of the FreqFLD without the FreqMoE and
    (c) the prediction results with the FreqMoE integrated. It can be seen that by integrating FreqMoE, the model can learn more effective facial structural information, thereby improving performance.}
\label{pregrd}
\end{figure*}

\begin{figure}[t]
\begin{center}
	\includegraphics[width=0.75\linewidth]{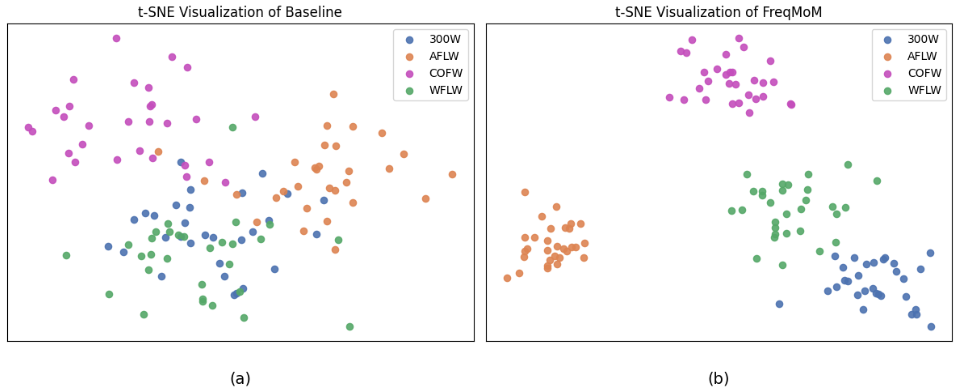}
	\end{center}
	\vspace{-2em}
	\caption{Comparison of t-SNE visualizations of features between the baseline and FreqMoM on latent layer. The results demonstrate that the proposed FreqMoM exhibits clear inter-dataset separation while remaining compact within each dataset.}
	\vspace{-1em}    
\label{tsne}
\end{figure}

\subsection{Self Evaluation}
The self evaluation provides comprehensive analyses of the design choices and practical behavior of the proposed FreqFLD. We examine the effects of the number of complexity experts and Top-$K$ routing, followed by evaluations of the FreqMoE and FreqMoM modules, and report time and memory efficiency.

\paragraph{\textbf{Evaluation of different numbers of complexity experts and Top-$K$ values}}
We investigated the effect of varying the number of activated experts. As shown in Table \ref{expert_num}, the model achieves the best performance among the evaluated expert configurations with 4 experts at TopK=1. Increasing the number of experts or activating multiple experts per sample does not further improve accuracy and even leads to performance degradation. This suggests that excessive experts introduce redundancy and insufficient expert specialization, while Top-1 selection encourages clear complexity-aware expert assignment, which is more suitable for FLD.

\paragraph{\textbf{Evaluation on FreqMoE}}
We analyze the dataset-level routing behavior of the proposed FreqMoE by measuring expert utilization induced by the routing function. As shown in Fig.~\ref{experts}, distinct expert utilization patterns emerge across different datasets. Samples from relatively less challenging datasets (e.g., 300W) are predominantly routed to lightweight experts, whereas more complex datasets such as WFLW and COFW exhibit higher activation frequencies of higher-capacity experts. These results indicate that FreqMoE adaptively allocates model capacity according to dataset-specific facial complexity. In addition, compared with the variant without FreqMoE (Fig.~\ref{pregrd}(b)), the model integrating FreqMoE (Fig.~\ref{pregrd}(c)) achieves more accurate and stable FLD performance, particularly under challenging conditions.

\paragraph{\textbf{Evaluation on FreqMoM}}
We evaluate FreqMoM on the 300W challenging subset. Starting from the Trans+FreqMoE baseline, we separately introduce the HFPB and LFPB. As shown in Table~\ref{tab_sensi}, HFPB reduces $\rm NME_{io}$ from 4.60 to 4.56, indicating improved modeling of fine-grained details, while LFPB achieves an $\rm NME_{io}$ of 4.58, reflecting its effectiveness in capturing global facial structure. When both branches are jointly integrated via FreqMoM, the model achieves the best performance with an $\rm NME_{io}$ of 4.52, highlighting the complementary roles of high- and low-frequency cues. 
In addition, Fig.~\ref{tsne} presents the t-SNE visualizations of feature representations learned by the baseline and FreqMoM. The baseline features in Fig.~\ref{tsne}(a) are highly mixed across different datasets, indicating limited discrimination under heterogeneous data distributions. By contrast, FreqMoM produces more compact intra-dataset clusters and clearer inter-dataset separation, as shown in Fig.~\ref{tsne}(b). This demonstrates that frequency modulation improves the discriminability of learned representations and helps alleviate feature conflicts in the All-in-One FLD setting.
These results can be attributed to the fact that high-frequency features primarily enhance sensitivity to fine-grained landmark details, while low-frequency features contribute to more stable modeling of global facial geometry, and their combination enables more geometry-sensitive FLD under the All-in-One paradigm.

\section{CONCLUSION}
In the All-in-One FLD task, learning a unified model that generalizes across heterogeneous datasets remains challenging. 
To address this issue, we present FreqFLD, an All-in-One framework that incorporates FreqMoM and FreqMoE to alleviate feature conflicts arising from heterogeneous facial data. Specifically, FreqMoM introduces explicit frequency modulation to disentangle low- and high-frequency facial cues, achieving balanced modeling of global structure and local details across heterogeneous datasets. Moreover, FreqMoE explicitly models the complexity of facial samples and adaptively routes them to experts with appropriate representational capacity, enabling robust handling of large variations in pose, expression, and appearance across datasets. In addition, the FreqCR loss is introduced to stabilize expert assignment and mitigate expert imbalance, thereby promoting consistent and robust expert specialization under diverse facial scenarios. By jointly leveraging FreqMoM, FreqMoE and FreqCR, the proposed FreqFLD alleviates  gradient conflicts in multi-dataset training, achieving comparable performance across multiple face alignment benchmarks. In the future, we will explore more lightweight frequency-aware expert designs to further reduce computational overhead while preserving the robustness of All-in-One FLD.

\section*{Acknowledgments}
\par{This work is supported by the National Natural Science Foundation of China (Grant NOs. 62476172, 62571555, 62576192 and 62576257), and the Natural Science Foundation of Hubei Province (2024AFB992).}


\bibliography{IEEEexample}

\end{document}